# Towards Discriminative and Interpretable Network: Deep Convolutional Decision Jungle

Seungryul Baek[1]
s.baek15@imperial.ac.uk

Kwang In Kim[2]
k.kim@bath.ac.uk

Tae-Kyun Kim[1]
tk.kim@imperial.ac.uk

[1] Imperial College London,
    London, UK

[2] University of Bath,
    Bath, UK

## Abstract

We propose a novel method called deep convolutional decision jungle (CDJ) and its learning algorithm for image classification. While the CDJ maintains the structure of the standard convolutional neural networks (CNNs) with multiple layers of multiple *response maps*, it aims at making intermediate representations more discriminative and interpretable, which in turn improves the classification accuracy of the network. During testing, each response map (or *node*) in both the convolutional and fully-connected layers selectively respond to class labels such that each data sample in effect travels via a specific *soft* route of those activated nodes. Therefore, CDJ's operational characteristics resemble those of the decision jungle. Compared to standard CNNs, the method embeds the benefits of using data-dependent discriminative functions, which better handles multi-modal/heterogeneous data. The network is learnt by combining the conventional softmax loss at the last layer and the proposed routing losses in each layer. The routing loss, as used in growing decision trees, measures the purity of data activation according to the class label distribution. Experiments on four real-world learning problems show that our method improves existing CNN architectures in the respective of both accuracy and interpretability.

## 1 Introduction

Random forests (RF) have been widely used for various computer vision problems, particularly on classification. RFs exhibit multiple inherent benefits from its tree structure: a complex classification problem is tackled in a hierarchical divide-conquer manner, and their training and testing steps are computationally efficient. Its structural diversity also motivates an ensemble learning. Convolutional neural networks (CNNs) have been proven to achieve state-of-the-art results in diverse problems. Its convolutional layers learn powerful features and fully connected layers and the last soft-max layer serve as a classifier. Recently, several works [3, 4, 9, 10, 14, 19, 23, 30, 32] have attempted combining the two methods (*e.g.* see Fig. 1a), for exploiting hierarchical tree-structures [4, 13, 14, 28], tackling multimodal data [30], clustering and learning modular networks per cluster [6, 19, 31, 32], and accelerating trainig and testing [9]. Also relevant to this study is to encourage sparsity in representation for regularization and efficiency in memory/time [3, 10, 12, 23]. Our method can be seen as a new regularizer that enforces sparsity per data point in testing time. The prior-works aforementioned reported improved accuracy; however, there is room to improve especially in the following aspects:

In the adopted binary tree structure [4, 9, 14, 30], once a data sample goes in a wrong path, it cannot be recovered *i.e.* they overfit. Soft partitioning helps relieve the issue to a certain degree, however, exponentially growing recursive binary splits [9, 30] remains as an issue. The decision jungle [24] type of algorithm becomes a natural extension for solving both and has shown good generalization ability. In another aspect, most existing methods for combining trees and CNNs [9, 30] require additional model parameters: The split process is often performed by a separate routing network [30]. Often the network layers need to be shallower to solve the explosion of the number of model parameters [9]. To summarize, it is challenging to apply such methods to existing large/deep networks [16, 26, 27]. [14] applied the tree structure using big networks; however tree structure exists only in the last fully connected layers. Our aim is to propose a novel method that applies *class entropy* or *purity* adopted from RFs to existing CNNs for more discriminative and interpretable representations in earlier layers and thus higher classification accuracy. The proposed architecture offers the following benefits:

**Class-wise purity in early layers**: Our method learns a convolutional neural network (CNN) with routing loss per layer. The proposed loss helps purify response maps, which we will call 'nodes', in all intermediate layers of CNNs. The response maps in each layer are *pushed on or off* conditionally on the input vectors such that each response map is dedicated to certain classes (ideally a class) than all. Note that the response maps for each data point take continuous values, thus this operation is treated as 'soft' routing. We observe that the supervised purification by class labels leads to more discriminative intermediate representations and higher accuracies.

**Data-dependent dynamic activation** by the decision jungle structure: Combination of CNNs and RFs is to improve the capacity of the given CNNs [32] thanks to its data-dependent discriminative functions. Compared to binary decision trees, the sample's paths are recoverable in later layers in the fully connected decision jungle structure. The decision jungle has shown improved generalization over binary trees. Also, such a structure makes the method be more flexible (applicable to any existing CNNs not altering their original architectures) and memory efficient than the binary tree structures [24] by sharing multiple branches.

**More Interpretable CNNs**: Several methods have been introduced [1, 2, 20, 34] to better understand intermediate representations learnt in CNNs. [1] propose a linear classifier as a probe to observe the behavior of intermediate layers. They use the probe for the trained CNNs and have shown that the entropy is a good measure for information contained in each layer. Similar to this, we evaluate each layer's discriminant power by the entropy measure. Further, we have an explicit mechanism to reduce the entropy in each layer by the gradient optimization. [2] propose a procedure for quantifying hidden layers' *interpretabilty* based on the alignment between individual hidden units and a set of semantic concepts evaluated on datasets. In the experiments (see Table 3), we demonstrate that our algorithm also significantly improves the interpretability from the baseline convolutional network architectures.

**Without additional routing parameters at testing**: The proposed architecture learns routing using additional routing path (the right path of Fig. 1(b)) during the training phase. At testing, this path is not used, which keeps the complexity of the model same as the original CNN models. Existing methods [9] require explicitly building routing networks at both train/test phases or change of the network structure to use binary routing [30] where the number of parameters increases exponentially (see Fig. 1(a)).

**Encoding auxiliary information**: Intermediate layers can be further purified by auxiliary labels (e.g. super-class labels or any other privileged information) in addition to the class labels. Experiments demonstrate significant accuracy improvements when using such auxiliary information.

## 2 Related work

**CNNs meet tree structures.** One of objectives in combining trees and CNNs is to learn both feature representations of input data and tree-structure classifiers conditioned on input data, in a joint manner [4, 9, 14, 30]. Previous methods can be categorized based on their operational characteristics as shown in Table 1: tree split structures in [4, 14] are embedded in the fully connected (FC) layers [14] rather than the convolutional (Conv) layers, which provide the final prediction as a classifier.

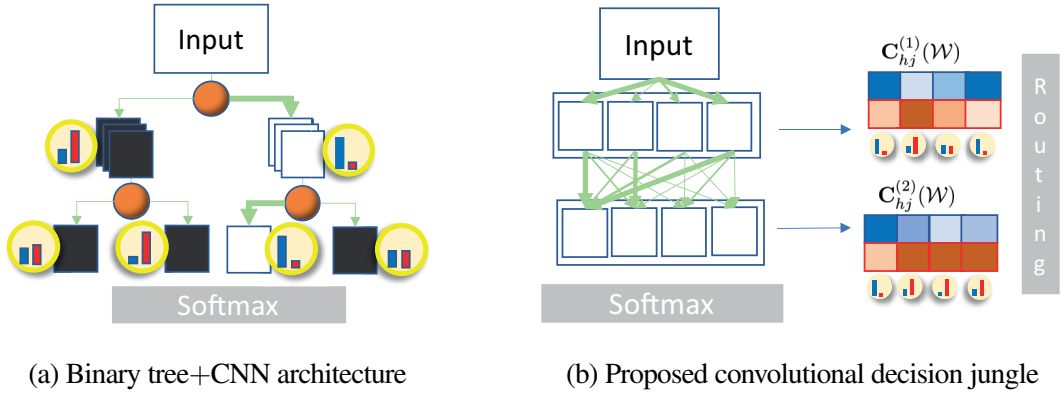

(a) Binary tree+CNN architecture      (b) Proposed convolutional decision jungle

Figure 1: Schematic diagram of the proposed convolutional decision jungle (b) compared to the binary tree+CNN architecture of [30] (a). Green arrows represent the routing where the degrees of *soft routing* are represented by their thickness. In (a), *hard* binary decisions are made by external routing networks (orange balls). The class conditional distributions are displayed as a histogram within each node. The black response maps at each node denotes the turned-off map by the binary routing, as proposed in [30]. Note that the number of nodes in each layer grows exponentially in (a). In (b), *soft* routing decisions are implicitly made by maximizing the class purity during training (Eq. 2). The class conditional distributions are obtained by applying the nonlinear activation functions (ReLU) and the average pooling to each intermediate layers' response map (displayed as blue arrows).

Table 1: Comparison to previous works that combine decision trees and CNN networks. Our algorithm does not require external routing networks, applicable to large-scale networks and straightforwardly incorporate auxiliary information.

|  | [4] | [14] | [30] | [9] | Ours |
|---|---|---|---|---|---|
| Binary hard split | ✓ | ✓ | ✓ | | |
| Multiway soft split | | | | ✓ | ✓ |
| Split structure in Convolutional layers | | | ✓ | ✓ | ✓ |
| Split structure in Fully Connected layers | ✓ | ✓ | | | ✓ |
| Without additional routing parameters at testing | | | | | ✓ |
| Scales to big networks | | ✓ | | ✓ | ✓ |
| Exploits auxiliary labels | | | | | ✓ |

In contrast, in [9, 30], split structures are embedded in the convolutional layers. They discover hidden modalities of data, *e.g.* face poses in [30], super-classes in [9], from the early layers. [9] uses continuous weights rather than discrete weights, thus their method can represent multiple *soft* routes than *hard* binary splits as in [4, 14, 30], However, the work has demonstrated routing only to 2-3 splits. A major difficulty in applying [9] to the fully connected networks such as decision jungle [24], where a node routes to all the nodes in the next layer, is in need of additional routing network parameters: Parameters increase as the number of routes increases. Furthermore, [4, 30] use small network architectures, while the methods in [9, 14] are applied to recently-proposed CNN architectures (AlexNet [16], VGG-16 Net [26], and GoogleNet [27]). Besides, the concept of using tree structures or conditional activation has been shown to improve CNN efficiency [3, 10].

Compared to the previous works [4, 9, 14, 30], our method does not require additional routing parameters, while enabling multiple soft routing and being applied to existing big architectures [9]. The experiments using three image classification and face verification benchmarks demonstrate improved accuracy (Sec. 4). When using additional label information, the method further improves accuracy via a more explicit network routing.

**Adding clustering loss to CNNs.** Though not explicitly using tree structures, relevant is a group of work that learns CNNs while performing hierarchical clustering [19, 23, 32]. The method in [32] iteratively performs feature representation learning and data clustering to better cluster data samples, while the method in [19] introduces a clustering loss to help classification. The concept of mixture of experts is introduced in the LSTM architecture [23] to improve its model capacity for language modeling. Set-based loss functions [22, 29] have also been proposed to encourage optimizing the intra-class/inter-class data variation, in addition to the softmax function. Compared to these existing methods, our method captures data separation from the early layers using class labels rather than exploiting sample distances in an unsupervised way.

# 3 Deep convolutional decision jungle

The proposed deep convolutional decision jungle (CDJ) adopts the conventional convolutional neural network (CNN) architecture. Given a fixed network topology (i.e. the number and size of layers), the operation of a CNN is prescribed by a set of weight vectors (or convolution filters) $\mathcal{W} = \{\mathbf{w}_{j,k}^{(l)}\}$:[1] the filter $\mathbf{w}_{j,k}^{(l)}$ at the $l$-th layer is convoluted with the $j$-th output (or *response*) map of the $(l-1)$-th layer $\mathbf{R}_j^{(l-1)}$ to generate the $l$-th layer's $k$-th response map $\mathbf{R}_k^{(l)}$:

$$\mathbf{R}_k^{(l)}(\mathbf{x}_i) = \sum_{j=1}^{R_l} \psi(\mathbf{R}_j^{(l-1)}(\mathbf{x}_i)) * \mathbf{w}_{j,k}^{(l)}, \tag{1}$$

where $R_l$ is the number of response maps in the $l$-th layer, $*$ denotes the convolution operation, and $\psi$ represents the non-linear activation and max-pooling operations. We use rectified linear units (ReLUs) while other activation functions can also be applied.

Our CDJ is instantiated by interpreting the response maps and convolution filters, respectively, as decision nodes and edges joining pairs of nodes in decision jungle (Fig. 1b). In traditional decision trees and jungles, an input $\mathbf{x}$ at a node is exclusively *routed* to a single child node. In our CDJ, a *soft decision* is made at each node $\mathbf{R}_k^{(l)}$ instead. If the filter $\mathbf{w}_{j,k}^{(l)}$ produces a non-zero response map $\mathbf{R}_k^{(l)}$, $\mathbf{x}$ is interpreted as being softly routed to the node $\mathbf{R}_k^{(l)}$.

## 3.1 Training deep convolutional decision jungles

Suppose we are given a set of training data points $\mathcal{X} = \{(\mathbf{x}_i, y_i)\}_{i=1}^N \subset \mathbb{R}^D \times \mathbb{R}$ with $D$ being the dimensionality of the input space. For a $C$-class classification problem, $y_i \in \{1, \ldots, C\}$. Training the CDJ corresponds to identifying the optimal set of convolutional filter parameters $\mathcal{W}^*$ achieved by minimizing the energy function that combines the standard training cost, a new *routing cost*, and a balancing cost:

$$\mathcal{C}(\mathcal{W}) = \sum_{i=1}^N \left[ \underbrace{\mathcal{C}_S(\mathbf{x}_i, y_i; \mathcal{W})}_{\text{training cost}} + \lambda_1 \underbrace{\sum_{l=1}^L \mathcal{C}_R^{(l)} \left( \mathbf{x}_i; \{\mathbf{R}_j^{(l)}(\mathbf{x}_i; \mathcal{W})\}_{i \in [1,N], j \in [1,R^l]} \right)}_{\text{routing cost}} \right.$$

$$\left. + \lambda_2 \underbrace{\sum_{l=1}^L \mathcal{C}_C^{(l)} \left( \mathbf{x}_i; \{\mathbf{R}_j^{(l)}(\mathbf{x}_i; \mathcal{W})\}_{i \in [1,N], j \in [1,R^l]} \right)}_{\text{label balancing cost}} \right], \tag{2}$$

---

[1]The weight vectors of the fully-connected layers are regarded as convolution filters of size 1.

where $\lambda_1, \lambda_2 \geq 0$ control the contributions of the routing cost and label balancing cost over the training cost, and $L$ is the number of layers in CDJ. We use the softmax loss for the training cost functional $\mathcal{C}_S$:

$$\mathcal{C}_S(\mathbf{x}_i, y_i; \mathcal{W}) = -\ln\left(\frac{e^{\mathbf{R}_{y_i}^{(L)}(\mathbf{x}_i; \mathcal{W})}}{\sum_{j=1}^{C} e^{\mathbf{R}_j^{(L)}(\mathbf{x}_i; \mathcal{W})}}\right). \tag{3}$$

With the goal of embedding decision jungle into a CNN architecture, we design our routing loss as a differentiable approximation of the class purity criteria used in decision forests and jungles [24]: Our routing loss enforces non-homogeneous responses according to inputs' class labels (*i.e.* each response is *dominated* by some classes) at the $l$-th layer:

$$\mathcal{C}_R^{(l)}\left(\mathbf{x}_i; \{\mathbf{R}_j^{(l)}(\mathbf{x}_i; \mathcal{W})\}_{i=1,\ldots,N, j=1,\ldots,R^l}\right) = -\frac{1}{R_l \cdot C} \sum_{j=1}^{R_l} \sum_{h=1}^{C} \left(\mathbf{C}_{hj}^{(l)}(\mathcal{W}) - \frac{1}{C} \sum_{h'=1}^{C} \mathbf{C}_{h'j}^{(l)}(\mathcal{W})\right)^2, \tag{4}$$

$$\mathbf{C}_{hj}^{(l)}(\mathcal{W}) = \sum_{i=1}^{N} a_{ij}^{(l)} \delta(y_i, h), \tag{5}$$

where $\delta(y_i, h) = 1$ if $y_i = h$ and $\delta(y_i, h) = 0$, otherwise. The scalar response $a_{ij}^{(l)}$ is obtained from $\mathbf{R}_j^{(l)}(\mathbf{x}_i; \mathcal{W}))$ via two layers that are differentiable: non-linear activation (ReLU) layer, followed by the average pooling layer is applied to the $\mathbf{R}_j^{(l)}(\mathbf{x}_i; \mathcal{W})$). The joint distribution $\mathbf{C}_{hj}^{(l)}(\mathcal{W})$ is a two-dimensional array storing the relative frequencies of patterns activated at each pair of response $j$ and class index $h$.

The interpretation of the routing loss as the class purity measure follows straightforwardly from its usage in decision trees (reducing *entropy*). However, naïvely applying this loss to training CNNs tends to generate degenerate solutions: The optimized solutions *disable* some classes by allocating (near-)zero probabilities to selected columns of (corresponding to specific classes) $\mathbf{C}_{hj}^{(l)}(\mathcal{W})$. This still contributes to reducing the overall entropy, but leads to poor generalization. Fig. 2b illustrates this problem with examples. Fig. 2e is the desirable solution compared to Fig. 2d.

To relieve the problem, the *Label balancing cost* is added to the optimization of Eq. 2. The cost is proposed to enforce the responses of each class label to have the similar scales to prevent from vanishing certain labels in the distribution of $\mathbf{C}^{(l)}$ at each layer:

$$\mathcal{C}_C^{(l)}\left(\mathbf{x}_i; \{\mathbf{R}_j^{(l)}(\mathbf{x}_i; \mathcal{W})\}_{i=1,\ldots,N, j=1,\ldots,R^l}\right) = \frac{1}{C} \sum_{h=1}^{C} \left(\sum_{j=1}^{R_l} \mathbf{C}_{hj}^{(l)}(\mathcal{W}) - \frac{1}{C} \sum_{h=1}^{C} \sum_{j=1}^{R_l} \mathbf{C}_{hj}^{(l)}(\mathcal{W})\right)^2. \tag{6}$$

The definition of routing loss and label balancing cost is consistent with our interpretation of CNN response maps as nodes in decision jungle. All three terms in the loss function (Eq. 2) are differentiable with respect to $\mathcal{W}$, and enable gradient descent-type optimization, e.g. stochastic gradient descent (SGD). We adopt mini-batch optimization instead of batch-optimizing the cost-functional $\mathcal{C}$ (Eq. 2). This facilitates efficient training using GPUs. Algorithm 1 summarizes the CDJ training process. We further derive the gradient update rules for our new modules (Eqs. 4 and 6) and perform experiments on the model variants using *training with auxiliary labels* and *ensemble learning*.

---

**Algorithm 1:** CDJ training algorithm.

**Input**: Training data
$\mathcal{X} = \{(\mathbf{x}_i, y_i)\}_{i=1}^{N}$ and initial filters $\mathcal{W}$, the number of epochs $T$, and the size of mini-batch $N'$;

**Output**: Learned filters $\mathcal{W}$;

**for** $t=1$ **to** $T$ **do**
    **for** $n=1$ **to** $N'$ **do**
        Construct $\mathbf{C}^{(l)}(\mathcal{W})$ from $n$-th mini-batch (Eq. 5);
        Update $\mathcal{W}$ based on sample gradients (Eq. 7);
    **end**
**end**

---

**Gradient update rule.** The cost $\mathcal{C}(\mathcal{W})$ in Eq. 2 is decomposed into three terms as follows:

$$\frac{\partial \mathcal{C}(\mathcal{W})}{\partial \mathcal{W}} = \sum_{i=1}^{N} \left[ \frac{\partial \mathcal{C}_S(\mathbf{x}_i, y_i; \mathcal{W})}{\partial \mathcal{W}} + \lambda_1 \sum_{l=1}^{L} \frac{\partial \mathcal{C}_R^{(l)}\left(\mathbf{x}_i; \{\mathbf{R}_j^{(l)}(\mathbf{x}_i; \mathcal{W})\}_{i \in [1,N], j \in [1, R^l]}\right)}{\partial \mathcal{W}} \right.$$
$$\left. + \lambda_2 \sum_{l=1}^{L} \frac{\partial \mathcal{C}_C^{(l)}\left(\mathbf{x}_i; \{\mathbf{R}_j^{(l)}(\mathbf{x}_i; \mathcal{W})\}_{i \in [1,N], j \in [1, R^l]}\right)}{\partial \mathcal{W}} \right], \tag{7}$$

where $\frac{\partial \mathcal{C}_S}{\partial \mathcal{W}}$ can be derived same as the standard CNNs, $\frac{\partial \mathcal{C}_R^{(l)}}{\partial \mathcal{W}} = \frac{\partial \mathcal{C}_R^{(l)}}{\partial \mathbf{C}_{hj}^{(l)}(\mathcal{W})} \cdot \frac{\partial \mathbf{C}_j^{(l)}(\mathcal{W})}{\partial a_{ij}} \cdot \frac{\partial a_{ij}}{\partial \mathcal{W}}$, $\frac{\partial \mathcal{C}_C^{(l)}}{\partial \mathcal{W}} = \frac{\partial \mathcal{C}_C^{(l)}}{\partial \mathbf{C}_{hj}^{(l)}(\mathcal{W})} \cdot \frac{\partial \mathbf{C}_{hj}^{(l)}(\mathcal{W})}{\partial a_{ij}} \cdot \frac{\partial a_{ij}}{\partial \mathcal{W}}$ where $\frac{\partial a_{ij}}{\partial \mathcal{W}} = \frac{\partial a_{ij}}{\partial \mathbf{R}_j^{(l)}(\mathbf{x}_i; \mathcal{W})} \cdot \frac{\partial \mathbf{R}_j^{(l)}(\mathbf{x}_i; \mathcal{W})}{\partial \mathcal{W}}$ follows from the chain rule as in the standard CNNs, and $\frac{\partial \mathcal{C}_R^{(l)}}{\partial \mathbf{C}_{hj}^{(l)}(\mathcal{W})}$ and $\frac{\partial \mathcal{C}_C^{(l)}}{\partial \mathbf{C}_{hj}^{(l)}(\mathcal{W})}$ can be calculated via $\frac{\partial x^2}{\partial x} = 2x$ and

$$\frac{\partial \mathbf{C}_{hj}^{(l)}(\mathcal{W})}{\partial a_{ij}} = \begin{cases} 1 & \text{if } \delta(y_i, h) = 1 \\ 0 & \text{otherwise} \end{cases}$$

for each $i$-th sample $\mathbf{x}_i$.

**Training with auxiliary labels.** Our new routing loss can be defined for any variables $\{z_i\}_{i=1}^{N}$ that replace the class labels $\{y_i\}_{i=1}^{N}$. This enables a systematic way of exploiting auxiliary information when available. We demonstrate the effectiveness of this approach in our face verification experiment using the pose as auxiliary labels. The class labels $\{y_i\}_{i=1}^{N}$ are used for the soft-max loss. See Table 2.

**Ensemble learning.** As in tree-based algorithms, our method inherits implicit structural diversity. In the experiments, we learnt 7 CDJ networks, and formed up an ensemble by combining their class predictions. The ensembled accuracy comparison between ensembled CNNs and ensembled CDJs on *Oxford-IIIT Pet* is shown in the Table 3.

## 4 Experiments

**Setup.** We evaluate our convolutional decision jungle (CDJ) on three image classification datasets: *Oxford-IIIT Pet* [21], *CIFAR-100* [15], and *Caltech-101* [5]; and a face verification dataset: *Multi-PIE* [8]. The *Oxford-IIIT Pet* dataset consists of 7,349 images covering 37 different breeds of cats and dogs (e.g., *Shiba Inu* and *Yorkshire Terrier*). We adopt the training and test splits of 3,680 and 3,669 images respectively [21, 22]. The *CIFAR-100* dataset contains 60,000 natural images of 100 classes that are grouped into 20 superclasses. We use a 50,000 training and 10,000 testing splits following [6, 7, 18, 19]. The *Caltech-101* dataset contains 9,146 images of 101 object categories. 30 images are chosen from each category for training and the remaining images (with at most 50 per category) are used for testing [11]. The *Multi-PIE* (Session 1 subset) face verification dataset consists of photographs of 250 individuals taken at 20 different illumination levels and 15 poses ranging from $-90°$ to $+90°$. We proceed as per [30]: For training, all images (15 poses, 20 illumination levels) of the first 150 individuals are used, while for testing a frontal view with neutral illumination (ID07 entries) is used as the *gallery* image for each of the remaining 100 subjects. The rest of the images are used as *probes*. We use the responses of the last hidden layer of our trained network as features for cosine distance-based matching. For each dataset, we adopt the competitive network architecture: AlexNet [16] and VGG-16 [26] for *Oxford-IIIT Pet* and *Caltech-101*, respectively, and NiN [18] for *CIFAR-100*. Each network is then initialized based on the standard ImageNet classification dataset. For *Multi-PIE*, our primary comparison is performed with c-CNN forests [30] that embed a decision tree into a *small* CNN architecture (Sec. 2). To show the flexibility (easy to use larger-scale network) of our method compared to c-CNN, we adopt AlexNet as our baseline.

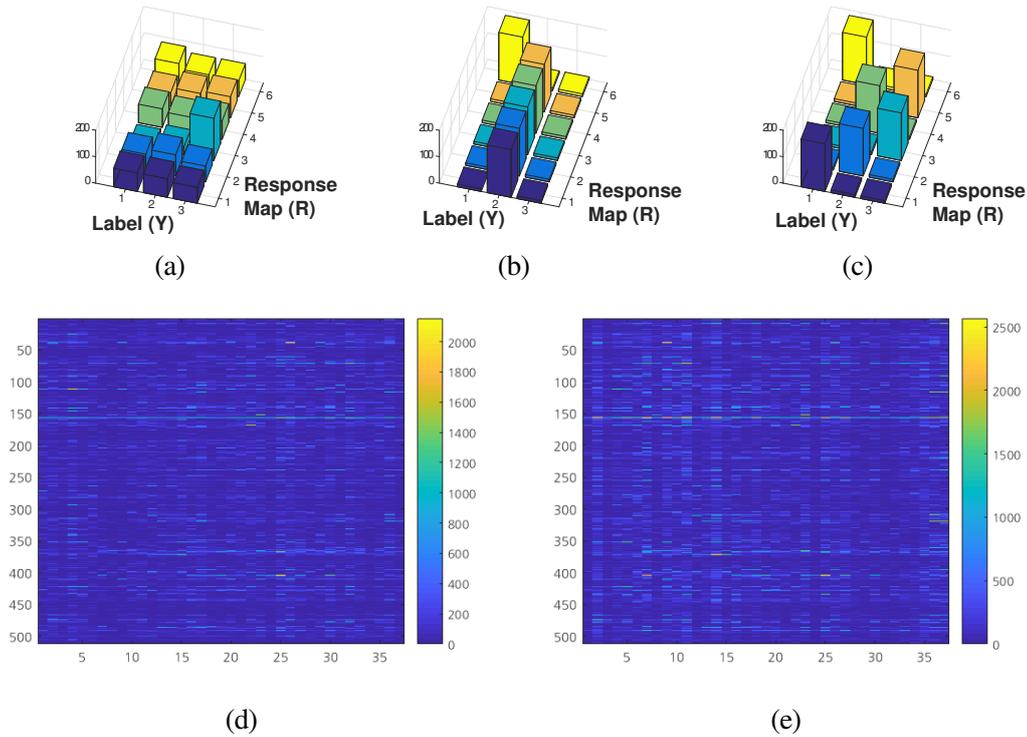

Figure 2: Examples response maps $\mathbf{C}_{hj}^{(l)}(\mathcal{W})$ in our CDJ: a toy example with $C=3$ and $R_l=6$. (a) $\mathbf{C}_{hj}^{(l)}(\mathcal{W})$ trained using only the softmax loss (the original CNN responses; $\lambda_1 = \lambda_2 = 0$); (b) and (c) the probability maps that minimize $\mathcal{C}$ (Eq. 2); Both (b) and (c) achieve smaller entropy values than (a) by suppressing a specific class ($Y=3$) in (b). However, the way of (b) is not desirable and we propose the *Label balancing constraint* in Eq. 2. The introduced *Label balancing constraint* makes the sum of each labels similar; thus the non-generate map as (c) could be obtained. (d) and (e) denotes the $\mathbf{C}_{hj}^{(l)}(\mathcal{W})$ from the 13-th convolutional layers (w/ 512 responses) from the vanilla VGG-16 and CDJ network trained on *Oxford-IIIT Pet* (w/ 37 classes). (e) has more peaks in each class labels compared to (d). This shows that filters in CDJ respond to input semantics more sensitively than those of vanilla networks.

The training parameters, including network topology, the mini-batch size, the number of training epochs, and dropout and local response normalization decisions are adopted from the respective state-of-the-art baseline: The mini-batch sizes are decided at 256, 32, and 100 for AlexNet, VGG-16, and NiN, respectively. The number of epochs are 60 for AlexNet and VGG-16, and 100 for NiN.

The learning rate is scheduled from $10^{-2}$ to $10^{-4}$ at every epoch for AlexNet, from $10^{-4}$ to $10^{-6}$ for VGG-16. Our learning rate for the NiN architecture is overall slightly smaller than [18], being fixed at 0.04 until the 80-th epoch, and reduced to 0.004 and 0.0004 after 80-th and 90-th epochs, respectively: This small learning rate ensures that the energy functional decreases constantly. For our new routing cost (Eq. 2), we use a larger mini-batch size of $10 \times C$ ($C$ being the number of classes) to ensure a balanced class distributions within each batch. Due to the *label balancing constraint* (Eq. 6), our routing cost is applicable only when the corresponding layer is larger than $C$. Therefore, we apply the routing loss from the second convolutional layer for AlexNet and VGG-16, and from the *cccp3* layer for NiN. The earlier layers—which do not use the routing loss—can be regarded as feature extractors of a decision jungle.

**Results.** Table 2(a)-(c) compare the results of our method with eleven state-of-the-art methods across three image classification problems. Overall, our algorithm constantly improves on each baseline with a large margin: Since we use the same network architecture as each baseline, these results can be attributed to the effectiveness of our new routing loss. More importantly, our method outperforms existing algorithms that incorporate special cost functions similarly to ours, including *squared hinge loss* applied to hidden layers (DSN [17]), *clustering cost* (DDN [19]), *set-based loss* of Magnet [22] that further uses data augmentation, and *label balancing cost* [11]. Our network attains routing from supervised class information rather than from unsupervised sample distances, as in these previous works, which helps learn more discriminative features.

Table 2(d) shows the results of different algorithms on *Multi-PIE* dataset. State-of-the-art performance on this dataset was achieved by c-CNN forests [30] which embed binary decision trees into a CNN architecture. To show the flexibility (easy to use larger-scale network) of our method compared to c-CNN, we adopt AlexNet as our baseline. As c-CNN forests doubles the number of nodes (or response maps) across each layer, it cannot be used to construct a *deep* network. This can pose severe application constraints: By simply training the deeper AlexNet, we already achieved much higher accuracy than c-CNN forests (Table 2(d)). In contrast, our CDJ facilitates large-scale, deep decision networks as it embeds computationally efficient and topologically flexible decision jungles into CNNs. The performance of the resulting CDJ is on par with the original AlexNet. However, CDJ inherits an important advantage of decision jungles: It is inherently tailored to heterogeneous feature modalities. By straightforwardly incorporating domain specific, auxiliary facial pose information into our routing cost, CDJ achieved significant improvement over AlexNet that already outperforms the state-of-the-art c-CNN forests (see Fig. 3(a) for the change in accuracy across different facial pose).

Figure 3(b-d) show the performance variation of the CDJ under hyper-parameter changes (*Oxford-IIIT Pet*): (b) and (c) show test accuracy variations with changes in epoch, and hyper-parameters $\lambda_1$ and $\lambda_2$, respectively; (d) shows the average entropy of each layer in CNNs and CDJs, respectively: Minimizing CNN's standard classification loss (softmax) tends to improve the class purity around the last layer. By explicitly enforcing it, our algorithm achieved low entropy even at the early layers. Note the strong correlation between the test accuracy (b) and (the inverse of) the average entropy at each layer (d) demonstrating the effectiveness of our new routing cost in improving generalization performance.

Finally, Table 3 shows the ensembled accuracies and average interpretability scores of the baseline CNNs and our method, respectively evaluated over all images and the response maps (*Oxford-IIIT Pet* dataset). The ensemble meaningfully improves the accuracy of CDJs over ensembled CNNs. Also, our algorithm achieved 19.81% and 16.18% increases of interpretability scores from AlexNet and VGG-16 baselines, respectively. The interpretability is defined based on the agreement of fine-grained ground-truth scene labels and the corresponding network responses as proposed by [2]. High interpretability facilitates the understanding of semantics developed within hidden units' of networks [2], which is an extremely important active problem in deep network research. Furthermore, the observed strong correlations between the accuracy and interpretability support [2]'s hypothesis that *disentangled representations* (as factorizations of the underlying problem structure) developed in the hidden layers are aligned with human-interpretable concepts.

Table 2: Results for three image classification and one face verification tasks: (a) *Oxford-IIIT Pet*, (b) *CIFAR-100*, (c) *Caltech-101*, and (d) *Multi-PIE*. 'Superclass info.' in (b) and 'Pose' column in (d) represent the use of auxiliary pose information, respectively.

| Method | Accuracy (%) |
|---|---|
| AlexNet [16] | 78.6 |
| GoogLeNet [27] | 88.7 |
| VGG-16 [26] | 88.8 |
| Magnet [22] | 89.4 |
| Ours (AlexNet) | 83.2 |
| Ours (VGG-16) | **90.5** |

(a)

| Method | Accuracy (%) |
|---|---|
| Maxout pooling [7] | 61.4 |
| NiN [18] | 64.3 |
| DSN [17] | 65.4 |
| DDN [19] | 68.3 |
| NiN+Superclass info. [6] | 68.8 |
| Ours (NiN) | 68.8 |
| Ours (NiN + Superclass info.) | **69.0** |

(b)

| Method | Accuracy (%) |
|---|---|
| Zeiler [33] | 86.5 |
| AlexNet [16] | 87.1 |
| LCN (AlexNet) [11] | 90.1 |
| VGG-16 [26] | 92.5 |
| LCN (VGG-16) [11] | 93.7 |
| Ours (AlexNet) | 90.5 |
| Ours (VGG-16) | **94.2** |

(c)

| Method | Accuracy | Pose |
|---|---|---|
| Fisher Vector [25] | 66.60 | ✗ |
| FIP 20 [35] | 67.87 | ✓ |
| FIP 40 [35] | 70.90 | ✓ |
| CNN 40 [30] | 70.81 | ✗ |
| c-CNN [30] | 73.54 | ✗ |
| c-CNN Forest [30] | 76.89 | ✗ |
| Baseline (AlexNet) | 88.98 | ✗ |
| Ours (AlexNet) | 88.04 | ✗ |
| Ours (AlexNet) + Pose label | **91.35** | ✓ |

(d)

Table 3: Ensembled accuracies and interpretability scores [2] on *Oxford-IIIT Pet* dataset.

| Method | Accuracy (%) | Interpretablity | Method | Accuracy (%) | Interpretablity |
|---|---|---|---|---|---|
| AlexNet [16] | 87.1 | 0.212 | Ours (AlexNet) | **89.5** | **0.254** |
| VGG-16 [26] | 92.3 | 0.272 | Ours (VGG-16) | **94.1** | **0.316** |

# 5 Conclusion

We presented a new convolutional neural network architecture and the corresponding training process called deep convolutional decision jungles, which shares the benefits of decision jungles and CNNs: class-wise purity at each node is maximized while end-to-end feature learning and classification are facilitated based on a single unified criteria. Compared to existing combinations of decision graphs and CNNs (e.g., c-CNN Forests [30]), our model offers higher flexibility in routing data and enables to exploit large-scale CNN architectures, thereby leading to enhanced performance and interpretability.

# Acknowledgements

This work is in part supported by EPSRC program grant FACER2VM (EP/N007743/1).

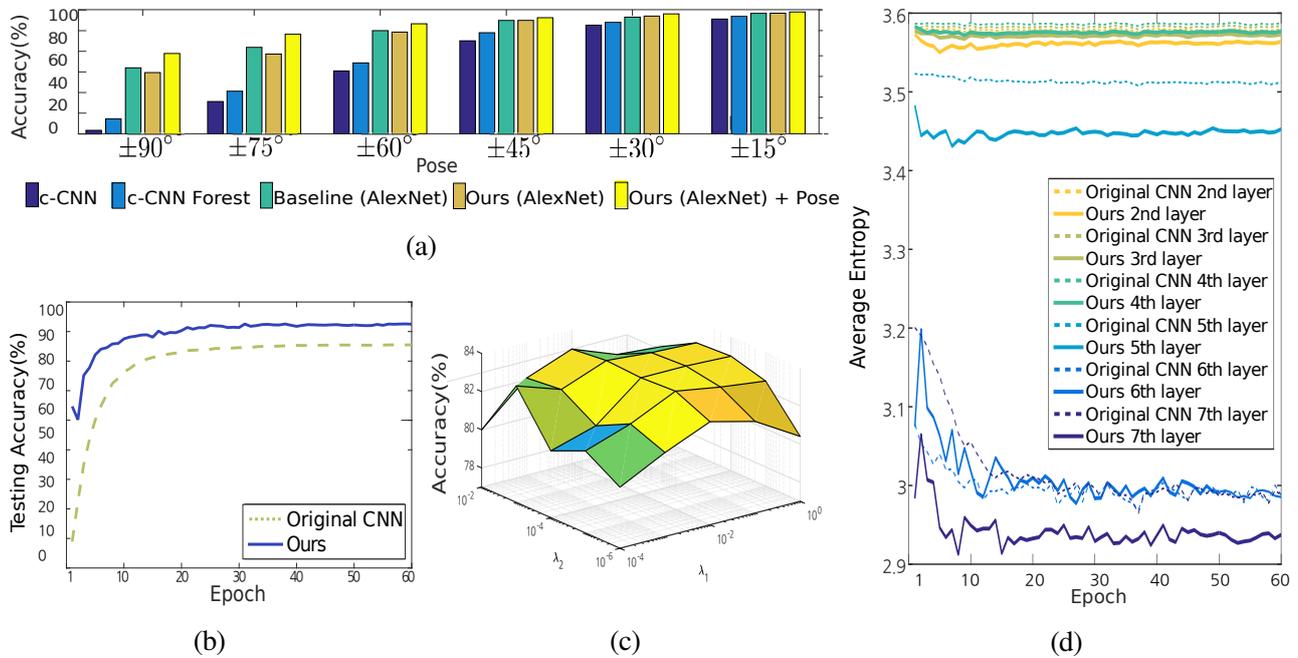

(a)

(b)                    (c)                    (d)

Figure 3: AlexNet results on *Multi-PIE* (a) and *Oxford-IIIT Pet* (b-d) datasets. (a) accuracy per human face over varying pose (out-of-plane horizontal rotation range); (b) test accuracy with respect to epoch: Our algorithm converges faster as guided by the new entropy energy; (c) test accuracy with varying hyper-parameters $\lambda_1$ and $\lambda_2$; The performance of our algorithm varies smoothly and predicatively; (d) average entropy of each layer in CNNs and CDJ: For both networks, entropy lowers toward the output layers. CDJ shows consistently lower entropy than CNN, especially at early layers.